\documentclass[a4paper]{article}

\usepackage{INTERSPEECH2021}
\usepackage{color,xcolor}

\title{Coreference Augmentation for Multi-Domain Task-Oriented \\Dialogue State Tracking}
\name{Ting Han$^{1*}$, Chongxuan Huang$^{2*}$, Wei Peng$^2$}
\address{
  $^1$University of Illinois at Chicago, Chicago, USA\\
  $^2$Artificial Intelligence Application Research Center, Huawei Technologies\\ Shenzhen, PRC}
\email{than24@uic.edu, \{huang.chongxuan, peng.wei1\}@huawei.com}

\begin{document}

\maketitle
\renewcommand{\thefootnote}{\fnsymbol{footnote}}
\footnotetext[1]{equal contribution}

\renewcommand{\thefootnote}{\arabic{footnote}}
\begin{abstract}
Dialogue State Tracking (DST), which is the process of inferring user goals by estimating belief states given the dialogue history, plays a critical role in task-oriented dialogue systems.  A coreference phenomenon observed in multi-turn conversations is not addressed by existing DST models, leading to sub-optimal performances. In this paper, we propose Coreference Dialogue State Tracker (CDST) that explicitly models the coreference feature. In particular, at each turn, the proposed model jointly predicts the coreferred domain-slot pair and extracts the coreference values from the dialogue context. Experimental results on MultiWOZ 2.1 dataset show that the proposed model achieves the state-of-the-art joint goal accuracy of 56.47\%.
\end{abstract}

\section{Introduction}
\begin{figure}[t]
  \centering
  \includegraphics[width=\linewidth, height=9cm]{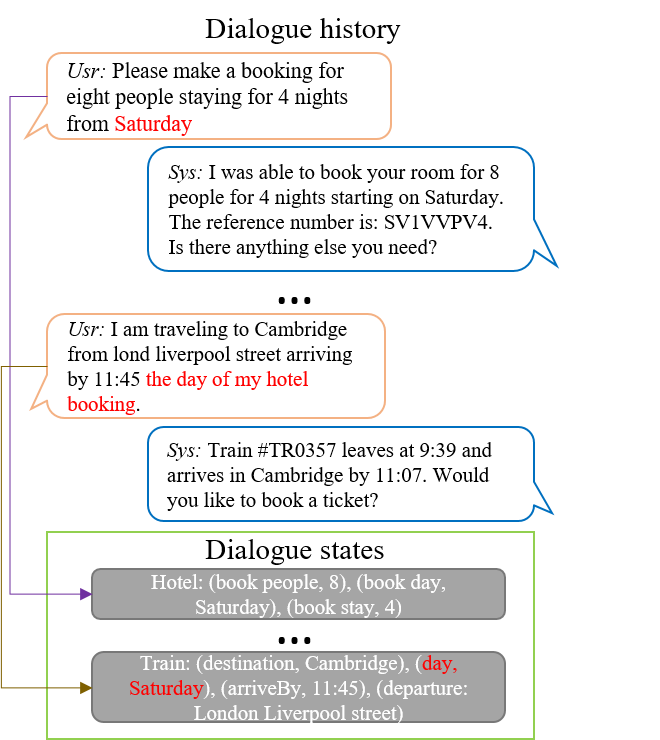}
  \caption{An example of dialogue state tracking in a conversation. The arrows connect user utterance and its paired dialogue state. The state tracker needs to track slot values mentioned by the user for all slots during the conversation.}
  \label{fig:dlgsample}
\end{figure}
Developing task-oriented dialogue systems has attracted interest from academia and industry due to its value in real-world applications. Dialogue state tracking (DST), one of the core functions of a task-oriented dialogue system, infers users' requirements by estimating the most probable belief states in the multi-run dialogue process. Given a user utterance, a DST model records user goals by filling a predefined slot set, as shown in Figure~\ref{fig:dlgsample}. The recorded state is updated at each turn and passed to downstream modules in the pipeline to generate a proper system response.

Traditional DST approaches rely on predefined ontology to produce belief states in terms of sets of slot-value pairs with the highest probability \cite{mrksic-etal-2017-neural, zhong-etal-2018-global, lee-etal-2019-sumbt, zhang-etal-2020-find}. It remains a challenge for these DST methods to address out-of-domain (OOD) dialogue states not accessible from the predefined ontology. Recent DST models \cite{wu-etal-2019-transferable, Chao2019, kim-etal-2020-efficient} resort to dialogue context to deal with the OOD problem mentioned above. Although progress has been made, they lack a mechanism to model the coreference features prevalent in human conversations, leading to sub-optimal performances. As illustrated in Figure~\ref{fig:dlgsample}, the slot value \textit{``Saturday"} (highlighted in red) mentioned in the previous turn does not explicitly appear in the current turn, in which only the coreferred term \textit{``the day of my hotel booking"} is present. It is a critical step to model the coreference features across dialogue turns in DST.

Several research works claim to improve the performance remarkably by accommodating coreference in the model. \cite{quan-etal-2019-gecor} show a significant performance enhancement through including coreference into training datasets. \cite{pan-etal-2019-improving} boost the response quality of the dialogue system by restoring incomplete utterances with coreference labels. \cite{su-etal-2019-improving} achieve a notable improvement to the baseline by recovering coreference information in the utterances. Recently, TripPy \cite{heck-etal-2020-trippy} implicitly models coreference relations among slots by a copy mechanism via which the coreference value for a particular slot is copied from other slots. However, the predefined schema adopted by TripPy for training coreference slot relationships is affected by annotation noise, thus preventing coreference feature learning.


\begin{figure*}
  \centering
  \includegraphics[width=\textwidth]{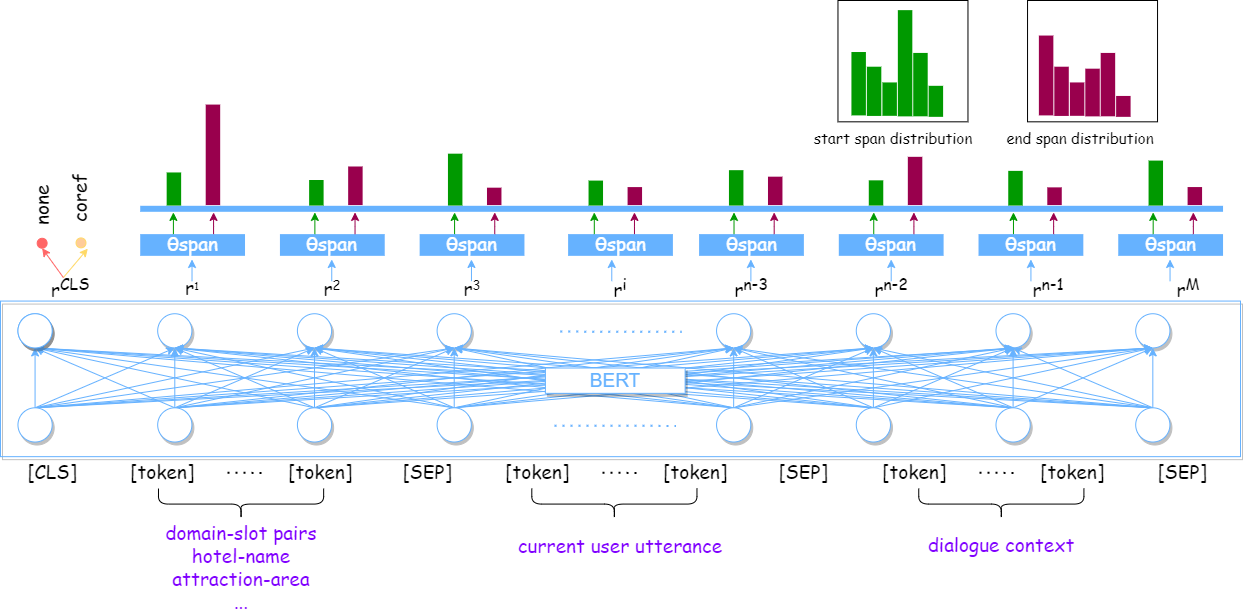}
  \caption{Network architecture of our proposed CDST. The input consists of domain-slot pairs, current user utterance and dialogue context, and the CDST outputs values for the domain-slot pairs if coreferences exist.}
  \label{fig:coref}
\end{figure*}


Instead of leveraging coreference-enriched datasets or predefined ontology mentioned above, we tackle the coreference problem in DST from another angle. Inspired by \cite{Chao2019}, we propose Coreference Dialogue State Tracker (CDST) to model the coreference feature explicitly. In particular, at each turn, the proposed model jointly predicts the coreferred domain-slot pair and extracts the coreference values from the dialogue context. Experimental results on MultiWOZ 2.1 \cite{eric-etal-2020-multiwoz} dataset demonstrate the effectiveness of CDST by achieving the state-of-the-art joint goal accuracy of 56.47\%. Additionally, we perform an empirical analysis to verify the necessity of the coreference solution.

The rest of the paper is organized as follows. In the next section, the proposed approach is presented. Section 3 provides experimental results and discussions followed by the conclusion of this paper.

\section{The proposed approach}
The network architecture of CDST is presented in Figure~\ref{fig:coref}. Denoting $X=\left\{\left(U_{1}, S_{1}\right), \ldots,\left(U_{T}, U_{T}\right)\right\}$ as the sequence of pairs which represents a dialogue of length $T$. $U_{t}$ and $S_{t}$ are user utterance and system utterance at turn $t$ respectively, and $DS=\{DS_{1}, {DS_{2}},..., DS_{N}\}$ is $N$ domain-slot pairs describing different subjects in a task-oriented dialogue system, for instance, \textit{Restaurant-name}. A restaurant name requested by a user during a conversation should be paired with the slot. CDST performs coreference dialogue state tracking by predicting coreference or ``none" value for each given slot $DS_{n}$ at each turn. As shown in Figure~\ref{fig:coref}, the contextual representations are extracted from sequential combinations of domain-slot pairs, the current user utterance and dialogue context via BERT \cite{devlin-etal-2019-bert}. The aggregated sentence representation, $r^{CLS}$ is used to differentiate the coreference slots from the none-coreferred ones. The token-level representations, $[ r^{1},..., r^{M}]$ are used to pinpoint coreference entities in the dialogue context of the input.

\subsection{Dialogue context encoder}
We adopt a pre-trained BERT as the contextual encoder. At each turn $t$, the $n_{th}$ domain-slot pairs $DS_{n}$, the current user utterance $U_{t}$ and dialogue history $C_{t}$ are concatenated as inputs to the encoder:
\begin{equation}
\begin{aligned}
 R_{tn} = {BERT([CLS] \oplus DS_{n} \oplus [SEP] \oplus} \\
  { U_{t} \oplus [SEP] \oplus C_{t} \oplus [SEP])}  \\
 \label{encode}
 \end{aligned}
\end{equation}
%
where [CLS] is the special token mandatory to be the first token of each input, and [SEP] is the special separator token. $C_{t} = (U_{1},S_{1}), ..., (U_{t-1},S_{t-1})$ denotes the dialogue context up to the previous turn $t-1$. The outputs of Equation (1) is the last hidden layer of BERT, i.e., $R_{tn} = [r_{tn}^{CLS}, r_{tn}^{1},..., r_{tn}^{M}]$, where $M$ denotes the total number of the input tokens, $r_{tn}^{CLS}$ is the aggregated representation of the entire input, which is later used by slot type classification. The remaining vectors $[r_{tn}^{1},..., r_{tn}^{M}]$ are the token-level representations for the input sequence and delivered to the coreference value predictions. The BERT is initialized from a pre-trained checkpoint and the parameters are further finetuned.

\subsection{Slot type classification}
One task of CDST is to determine, for each domain-slot pair $DS_n \in DS$, if the domain-slot is coreferred. It can be treated as a binary classification on a variable with values $\{coref, none\}$, indicating whether coreference exists for the given domain-slot pair. The classification takes $r_{tn}^{CLS}$ as the input and outputs the probability of being coreferred for the domain-slot pair $DS_n$ at the turn $t$.
\begin{equation}
\begin{aligned}
 P_{tn} = softmax(W_{n}^{c} \cdot r_{tn}^{CLS} + b_{n}^{c})
 \label{softmax}
 \end{aligned}
\end{equation}
where $W_{n}^{c}$ and $b_{n}^{c}$ are trainable parameters for the domain-slot pair $DS_n$.

\subsection{Coreference value retrieval}
Simultaneously, CDST retrieves coreference value from the given dialogue context through span prediction. For each domain-slot pair $DS_n \in DS$, a span prediction layer takes each token of $[r_{tn}^{1},..., r_{tn}^{M}]$ as the input and outputs the start and end positions for the token:
\begin{equation}
\begin{aligned}
 (start_{tn}^i, end_{tn}^i) = W_{n}^{span} \cdot r_{tn}^{i} + b_{n}^{span}
 \label{position}
 \end{aligned}
\end{equation}

\begin{equation}
\begin{aligned}
 P_{tn, start} = softmax(start_{tn})
 \label{start_prob}
 \end{aligned}
\end{equation}

\begin{equation}
\begin{aligned}
 P_{tn, end} = softmax(end_{tn})
 \label{end_prob}
 \end{aligned}
\end{equation}
where $W_{n}^{span}$ and $b_{n}^{span}$ are learnable parameters. The final span for the domain-slot pair $DS_n$ is obtained through an argmax function over the start and end positions of all tokens:
\begin{equation}
\begin{aligned}
 SPAN_{tn, start} = argmax(P_{tn,start})
 \label{start position}
 \end{aligned}
\end{equation}
\begin{equation}
\begin{aligned}
 SPAN_{tn, end} = argmax(P_{tn,end})
 \label{end position}
 \end{aligned}
\end{equation}


The slot type classification and coreference value retrieval coordinate each other to complete coreference dialogue state tracking. The retrieved coreference value is filled into the given domain-slot pair if it is classified as coreferred. Otherwise, the retrieved values are discarded.

\section{Experiments and results}

\begin{figure}[t]
  \centering
  \includegraphics[width=\linewidth]{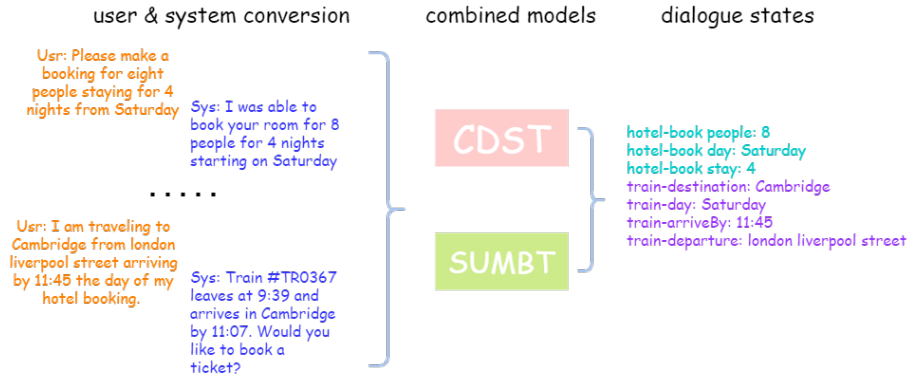}
  \caption{An illustration of the combination of CDST and SUMBT. The two models work jointly as a dialogue state tracker.}
  \label{fig:combine}
\end{figure}

We conduct three different experiments: (1) evaluating CDST solely on coreference; (2) jointly examining CDST along with SUMBT\footnote{We skip the details of SUMBT, which is accessible in the references.} \cite{lee-etal-2019-sumbt} on the purpose of a DST task, and a combination of two models is provided in Figure~\ref{fig:combine}; (3) adopting T5 \cite{2020t5} for comparisons and the empirical analysis with CDST. We use the joint goal accuracy (JGA) as the evaluation metric. 

\subsection{Dataset}

All three experiments are conducted on the MultiWOZ 2.1 dataset, which is a multi-domain task-oriented dialogue dataset with more than 10k dialogues spanning over seven different domains. There are 30 slots and five domains (train, restaurant, hotel, taxi, attraction) commonly used for DST tasks. The number of dialogues for the training, development, and test sets are 8,348, 1,000, and 1,000, respectively. The coreference data is from MultiWOZ 2.3 \cite{han2020multiwoz}, an improved dataset based on MultiWOZ 2.1 with extra coreference annotations. In total, 20.16\% of the dialogues contain coreference.

\subsection{Implementation details}
We adopt BERT-medium, uncased\footnote{https://github.com/google-research/bert} \cite{turc2019well} for CDST, and use a T5-base framework from Simple Transformers\footnote{https://github.com/ThilinaRajapakse/simpletransformers}. The training hyperparameters of the two models are listed in the Table~\ref{tab:paras}.
\begin{table}[th]
  \caption{Hyperparameters for BERT-medium and T5-base}
  \label{tab:paras}
  \centering
  \begin{tabular}{|c|c|c|}
    \hline
    \textbf{Model Hyperparameters}      & \textbf{BERT-Medium} & \textbf{T5-base}  \\
    \hline
    learning rate & $1 \times 10^{-4}$ &$1 \times 10^{-3}$ \\
    \hline
    max seq length & 512 & 196 \\
    \hline
    warmup ratio & 0.1 & 0.06 \\
    \hline
    train epoch & 10 & 20 \\
    \hline
    optimizer & ADAM & Adafactor \\
    \hline
    train batchsize & 2 & 16 \\
    \hline
  \end{tabular}
\end{table}

The loss function of the T5-base remains intact during training. Comparatively, the loss of CDST is computed through a summation of two loss components as follow:

\begin{equation}
\begin{aligned}
 \mathcal{L}_{total} = \beta \cdot \mathcal{L}_{slot\_type} + (1 - \beta) \cdot \mathcal{L}_{SPAN}
 \label{total_loss}
 \end{aligned}
\end{equation}
%
where $\mathcal{L}_{slot\_type}$ and $\mathcal{L}_{SPAN}$ are cross-entropy loss for the corresponding predictions. $\mathcal{L}_{SPAN}$ is the average loss of the span start and end losses. $\beta$ is a parameter that is empirically set to 0.8.

\begin{figure*}
  \centering
  \includegraphics[width=\textwidth, height=8cm]{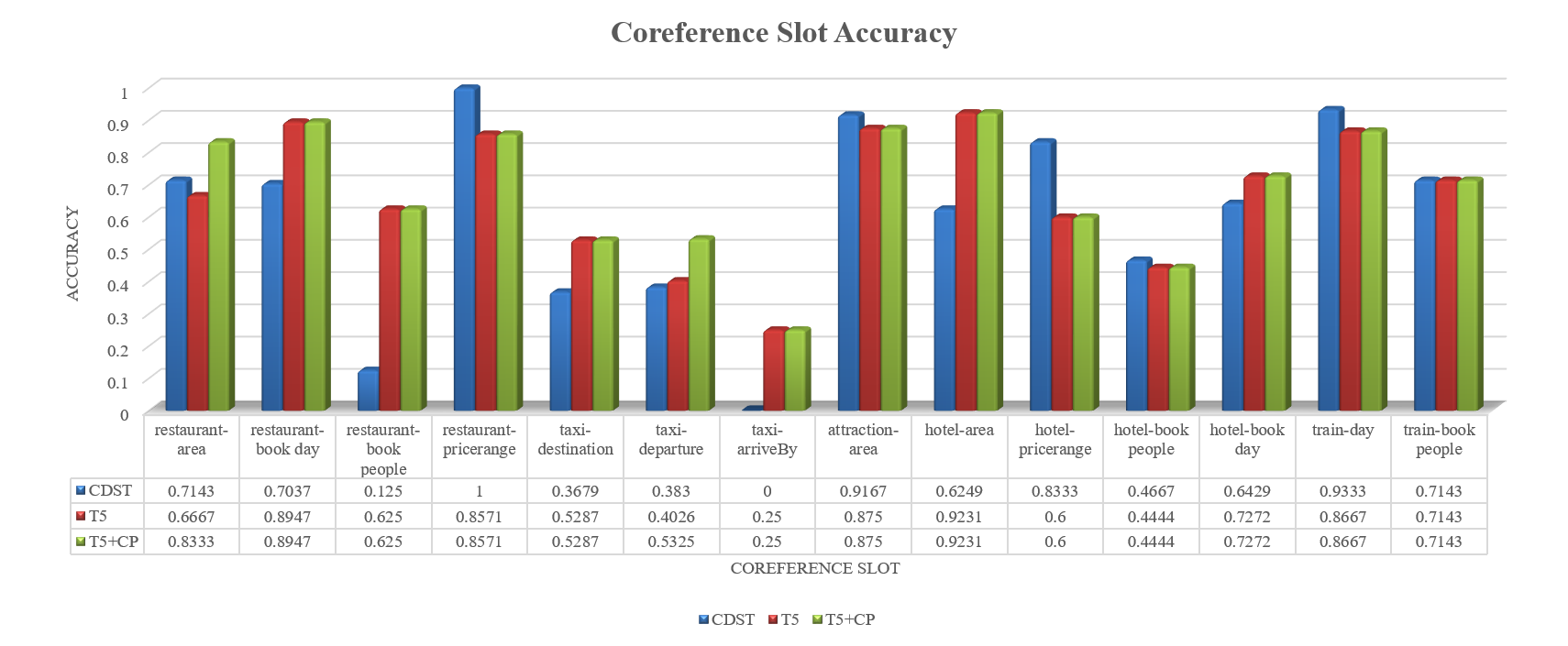}
  \caption{A comparison of the accuracy of coreference slots of CDST, T5 and T5+CP.}
  \label{fig:slot_rate1}
\end{figure*}

We finetune T5's encoder-decoder architecture to handle coreference modeling, which is treated as a question-answering (QA) task. The question includes the domain-slot pair concatenated with the current user utterance. We treat dialogue context as the passage in the QA task. T5 is trained using coreferred data samples to maximize the performance of the QA task. The predicted coreference value for a particular slot is used to update the slot. A rule-based mechanism is applied to merge results of T5 into SUMBT before evaluating the joint goal accuracy. 

\subsection{Results}

As shown in Table~\ref{tab:multiwoz_table}, SUMBT+CDST outperforms other dialogue state trackers by achieving the state-of-the-art joint goal accuracy (JGA). CDST adds 3.9\% of JGA to the performance of SUMBT, demonstrating its effectiveness in the DST task. Although the result of SUMBT+T5 is comparable to our proposed model, it contains twice as many parameters (220 million) as that of the CDST, putting T5 a less desirable option.

\begin{table}[th]
  \caption{DST results on MultiWOZ 2.1. The result of the original SUMBT is trained and evaluated on MultiWOZ 2.0, depicted with a notation `$^*$'. We further implement SUMBT with MultiWOZ 2.1 and the result is labelled by `$^+$'.}
  \label{tab:multiwoz_table}
  \centering
  \begin{tabular}{|c|c|}
    \hline
    \textbf{Models}      & \textbf{MultiWOZ 2.1}                \\
    \hline
    SUMBT \cite{lee-etal-2019-sumbt}    & 42.40\%$^*$        \\
    TRADE \cite{wu-etal-2019-transferable}   & 45.60\%                                \\
    DSTQA \cite{zhou2019multi} & 51.17\%                          \\
    
    DS-DST \cite{zhang-etal-2020-find} & 51.21\%                               \\
        SOM-DST \cite{kim-etal-2020-efficient} & 52.57\% \\
    DST-picklist \cite{zhang-etal-2020-find} & 53.50\% \\

    SST \cite{Chen_Lv_Wang_Zhu_Tan_Yu_2020} & 55.23\% \\
    TripPy \cite{heck-etal-2020-trippy} & 55.29\% \\
    \hline
    SUMBT & 52.57\%$^+$ \\
    SUMBT+T5 & 56.41\% \\
     SUMBT+CDST & \textbf{56.47\%} \\
    \hline
  \end{tabular}
\end{table}
\subsection{Empirical analysis and discussions}
Empirical analyses on the coreference results of T5 and CDST are summarized in Table~\ref{tab:coreference} and Figure~\ref{fig:slot_rate1}.


We perform an ablation study on coreference performance under various combinations for inputs. Table~\ref{tab:coreference} presents the performances of different input combinations among domain-slot pairs, current user utterance, and dialogue context. It can be observed that current user utterance (uttr.) and domain-slot pairs (aka slot.) contribute approximately 10\% of the joint goal accuracy (JGA) to the CDST, boosting the performance from 45.52\% to 55.84\%.  

\begin{table}[th]
  \caption{Coreference results of CDST and T5 in an ablation study. ``-uttr." means experimenting without including the current user utterance in the input. ``-slot." refers to an experiment without inputting domain-slot pairs. We do not record the score for ``T5-uttr.,-slot." because ``slot." has been treated as a question (a mandatory input) by T5 in this study.}
  \label{tab:coreference}
  \centering
  \begin{tabular}{|c|c|}
    \hline
    \textbf{Models}      & \textbf{Accuracy}                \\
    \hline
    T5+CP                  & 70.72\%                                \\
    T5 & 65.13\% \\
    T5-uttr.            & 58.55\%                          \\
    T5-uttr.,-slot.   & -  \\                  
    \hline
    \textbf{Models} & \textbf{JGA} \\
    \hline
    CDST & 55.84\% \\
    CDST-uttr. & 48.31\% \\
    CDST-uttr., -slot. & 45.52\% \\
    \hline
  \end{tabular}
\end{table}

In the ablation test, we evaluate T5 using accuracy as the performance metric instead of JGA because T5 alone can only handle single-turn dialogue data. It is confirmed that current user utterance (uttr.) and domain-slot pairs (slot.) also contribute significantly to T5. We further include ``coreference expression phrase" (``CP"), i.e., ``the day" in Figure~\ref{fig:dlgsample}, into the experiment to investigate their effects on the DST task when using T5.  As shown in Table~\ref{tab:coreference}, the performance of T5 is significantly enhanced to 70.72\% as a consequence.

In the Figure~\ref{fig:slot_rate1}, we record the coreference accuracy per slot for the top 3 performances in Table~\ref{tab:coreference}. In total, there are 14 domain-slot pairs with coreference. It can be observed that: 
\begin{itemize}
    \item Both T5 and CDST are insensitive to numbers leading to relatively lower accuracies for slots related to ``book people". The worst performance is for ``taxi-arriveBy" which is a time-related slot. 
    \item The poor accuracy of ``taxi-destination" and ``taxi-departure" indicates a lack of discriminating power for the model to handle departure and destination information concurrently. The slight improvement of the performance of T5+CP relating to ``taxi-departure" may be attributed to the information introduced by the ``coreference context phrase" (CP). 
    \item The inclusion of the ``coreference expression phrase" (CP) has only benefited two slots (``restaurant-area" and ``taxi-departure"). Further works are required to pinpoint the potential cause of this limitation. 
\end{itemize}

\section{Conclusion}

In this paper, we propose Coreference Dialogue State Tracker (CDST) to explicitly model coreference in the DST task. CDST directly retrieves coreferred slot values from the given dialogue context without a predefined ontology. The experimental results demonstrate the effectiveness of CDST by achieving the state-of-the-art joint goal accuracy. Furthermore, we conduct an empirical analysis to disclose the insights associated with coreference modeling in DST.





\bibliographystyle{IEEEtran}

\bibliography{mybib}


\end{document}